\documentclass[runningheads]{llncs}
\usepackage{graphicx}
\usepackage[ruled,vlined]{algorithm2e}
\usepackage{algorithmic}
\usepackage{tikz}
\usepackage{comment}
\usepackage{amsmath,amssymb} 
\usepackage{color}
\usepackage{ulem}
\usepackage[accsupp]{axessibility}  

\usepackage{makecell}
\usepackage{multirow}
\usepackage{booktabs}
\usepackage{wrapfig}

\usepackage[pagebackref,breaklinks,colorlinks]{hyperref}
\usepackage{breakcites}
\makeatletter
\newcommand{\printfnsymbol}[1]{%
\textsuperscript{\@fnsymbol{#1}}%
}

\begin{document}
\pagestyle{headings}
\mainmatter
\def\ECCVSubNumber{2667}  

\title{Improving the Reliability for Confidence Estimation}

\titlerunning{Improving the Reliability for Confidence
Estimation}
%
\author{Haoxuan Qu\inst{1}\thanks{Both authors contributed equally to the work.}
\and
Yanchao Li\inst{1}\printfnsymbol{1}
\and
Lin Geng Foo\inst{1}\index{Foo, Lin Geng}
\and
Jason Kuen\inst{2}
\and
Jiuxiang Gu\inst{2}
\and
Jun Liu\inst{1}\thanks{Corresponding Author}
}
\authorrunning{H. Qu et al.}
%
\institute{Singapore University of Technology and Design \\
\email{\{haoxuan\_qu, lingeng\_foo\}@mymail.sutd.edu.sg, \{yanchao\_li, jun\_liu\}@sutd.edu.sg} \and
Adobe Research\\
\email{\{kuen, jigu\}@adobe.com}}
\maketitle

\begin{abstract}

Confidence estimation, a task that aims to evaluate the trustworthiness of the model's prediction output during deployment, has received lots of research attention recently, due to its importance for the safe deployment of deep models. Previous works have outlined two important qualities that a reliable confidence estimation model should possess, i.e., the ability to perform well under label imbalance and the ability to handle various out-of-distribution data inputs. In this work, we propose a meta-learning framework that can simultaneously improve upon both qualities in a confidence estimation model. Specifically, we first construct virtual training and testing sets with some intentionally designed distribution differences between them. Our framework then uses the constructed sets to train the confidence estimation model through a \textit{virtual training and testing} scheme leading it to learn knowledge that generalizes to diverse distributions. We show the effectiveness of our framework on both monocular depth estimation and image classification.
\keywords{Confidence estimation, Meta-learning.}
\end{abstract}

\setlength\intextsep{0pt}
\section{Introduction}

With the continuous development of deep learning techniques, deep models are becoming increasingly accurate on various computer vision tasks, such as image classification \cite{krizhevsky2012imagenet} and monocular depth estimation \cite{lee2019big}. 
However, even highly accurate models might still commit errors \cite{amodei2016concrete,janai2020computer,floridi2019establishing}, and these errors can potentially lead to serious consequences, especially in safety-critical fields, such as nuclear power plant monitoring \cite{linda2009neural}, disease diagnosis \cite{sanz2014medical}, and self-driving vehicles \cite{shafaei2018uncertainty}.
Due to the severe implications of errors in these applications, it is crucial for us to be able to assess whether we can place confidence in the model predictions, before acting according to them.
Hence, the task of confidence estimation (also known as trustworthiness prediction), which aims to evaluate the confidence of the model's prediction during deployment, has received a lot of research attention recently \cite{jiang2018trust,corbiere2019addressing,luo2021learning}.

\begin{wrapfigure}[8]{r}{0.6\textwidth}
\centering
\includegraphics[width=0.5\textwidth]{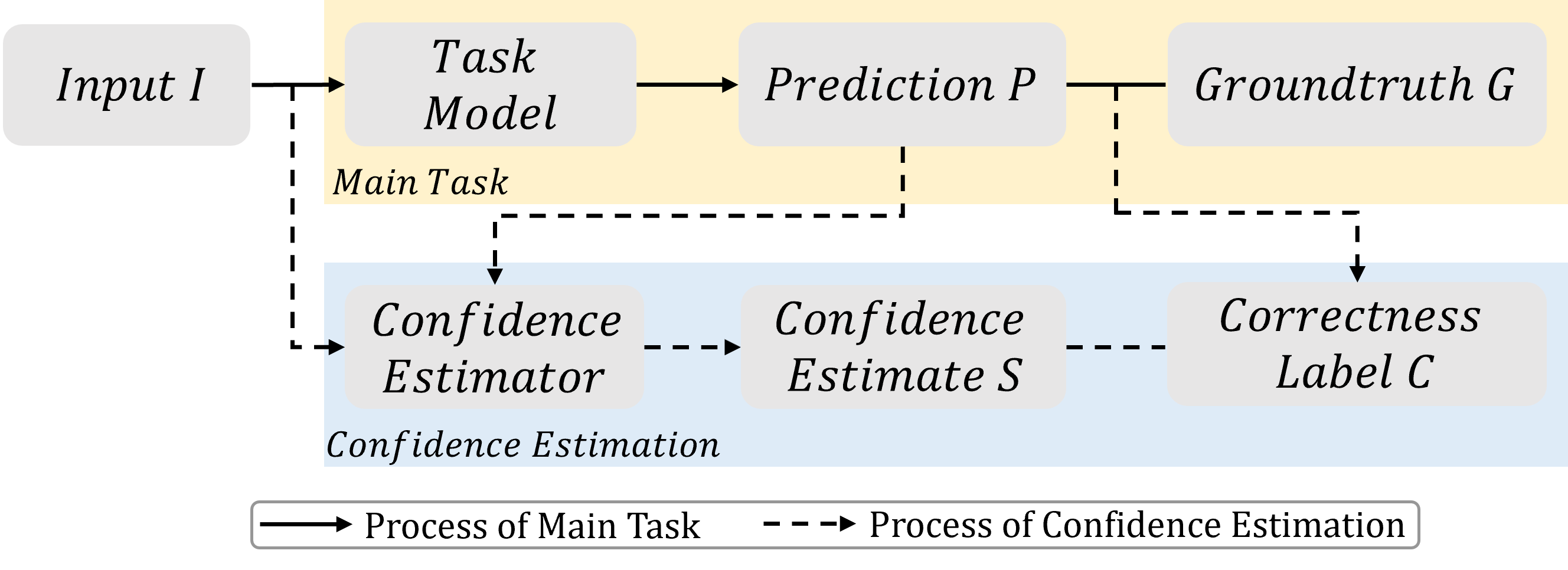}
\caption{\textbf{Illustration of confidence estimation. }}
\label{fig:illu}
\end{wrapfigure}

Specifically, in confidence estimation, we would like to compute the confidence estimate $S \in \{0,1 \}$ for a prediction $P$ made by a model regarding input $I$, where $S$ estimates if prediction $P$ is correct (1) or not (0).
In this paper, for clarity, the \textit{task model} refers to the deep model that produces predictions $P$ on the main task;
confidence estimation for $P$ is performed by a separate confidence estimation model, which we refer to as the \textit{confidence estimator}, as shown in Fig.~\ref{fig:illu}.
Many previous works \cite{corbiere2019addressing,luo2021learning,li2021identifying,yu21bmvc} have proposed to train such a confidence estimator to conduct confidence estimation more reliably.

Some recent works \cite{luo2021learning,li2021identifying} have noted that a reliable confidence estimator should \textit{perform well under label imbalance}.
This is because confidence estimators use the \textit{correctness of task model predictions ($C$) as labels}, which are often imbalanced.
As shown in Fig.~\ref{fig:illu}, correctness labels $C$ are produced by checking for consistency between predictions $P$ and ground truths $G$, where $C=1$ if $P$ is correct and $C=0$ otherwise. 
Thus, since many task models have achieved good performance on computer vision tasks (e.g., Small ConvNet \cite{jiang2018trust} achieves $> 99\%$ for MNIST \cite{lecun1998gradient} and VGG-16 \cite{SimonyanZ14a,liu2015very} achieves $> 93\%$ for CIFAR-10 \cite{krizhevsky2009learning} in image classification), there are often many more correct predictions (where $C=1$) than incorrect ones (where $C=0$), which leads to label imbalance for the confidence estimation task.
If this label imbalance is not accounted for during training, the confidence estimator is likely to be overly confident \cite{luo2021learning,li2021identifying} for incorrect predictions (where $C=0$), which is undesirable.

On the other hand, some other works \cite{mukhoti2020calibrating,tomani2021post} have suggested that the ability to handle \textit{out-of-distribution data inputs} ($I$) is important for confidence estimation.
Out-of-distribution data occurs due to distribution shifts in the data --
such data distribution shifts can occur within the same dataset \cite{matsuura2020domain}, but are generally more severe between different datasets,
e.g. between the training data from existing datasets and testing data received during deployment under real-world conditions.
If the confidence estimator does not learn to handle out-of-distribution inputs, it will tend to perform badly whenever an out-of-distribution input sample $I$ is fed into the task model, which affects its utility in practical applications.

In this paper, we aim to improve the reliability of our confidence estimator in terms of both the above-mentioned qualities; we improve its ability to tackle label imbalance of $C$, as well as to handle various out-of-distribution inputs $I$. 
Specifically, we observe that these qualities actually share a common point -- they are acquired when the confidence estimator learns to \textit{generalize to diverse distributions}.
If a confidence estimator learns knowledge that can generalize to diverse distributions, it will be able to tackle diverse correctness label ($C$) distributions, which includes distributions where $C=0$ is more common, and can thus better tackle the imbalanced label problem;
it will also be able to tackle diverse input ($I$) distributions, which improves performance on out-of-distribution data.
Based on this novel perspective, we propose to improve upon both of these qualities \textit{simultaneously through a unified framework}, that allows the confidence estimator to learn to generalize, and perform well on distributions that might be different from the distributions (of both $C$ and $I$) seen during training.
In order to achieve this, we incorporate \textit{meta-learning} into our framework.

Meta-learning, also known as ``\textit{learning to learn}'', allows us to train a model that can \textit{generalize well to different distributions}.
Specifically, in some meta-learning works \cite{finn2017model,li2018learning, guo2020learning,bai2021person30k,huang2021metasets,xu2022meta}, a \textit{virtual testing set} is used to mimic the testing conditions during training, so that even though training is mainly done on a \textit{virtual training set} consisting of training data, performance on the testing scenario is improved.
In our work, we construct our virtual testing sets such that they simulate various distributions that are different from the virtual training set, which will push our model to learn \textit{distribution-generalizable knowledge to perform well on diverse distributions}, instead of learning \textit{distribution-specific knowledge that only performs well on the training distribution}.
In particular, for our confidence estimator to learn distribution-generalizable knowledge and tackle diverse distributions of $C$ and $I$, we intentionally construct virtual training and testing sets that simulate the different distribution shifts of $C$ and $I$, and use them for meta-learning.

The contributions of our work are summarized as follows.
1) We propose a novel framework, which incorporates meta-learning to learn a \textit{confidence estimator} to produce confidence estimates more reliably.
2) By carefully constructing virtual training and testing sets that simulate the training and various testing scenarios, our framework can learn to generalize well to different correctness label distributions and input distributions.
3) We apply our framework upon state-of-the-art confidence estimation methods \cite{corbiere2019addressing,yu21bmvc} across various computer vision tasks, including image classification and monocular depth estimation, and achieve consistent performance enhancement throughout.

\section{Related Work}

\noindent\textbf{Confidence Estimation.}
Being an important task that helps determine whether a deep predictor's predictions can be trusted,
confidence estimation has been studied extensively across various computer vision tasks \cite{hendrycks17baseline,gal2016dropout,jiang2018trust,corbiere2019addressing,moon2020confidence,qiu2020detecting,luo2021learning,tsiligkaridis2021failure,chen2021detecting,mukhoti2020calibrating,li2021identifying,tomani2021post,yu21bmvc}. 
At the beginning, Hendrycks and Gimpel \cite{hendrycks17baseline} proposed Maximum Class Probability utilizing the classifier softmax distribution, Gal and Ghahramani \cite{gal2016dropout} proposed MCDropout from the perspective of uncertainty estimation, and Jiang et al. \cite{jiang2018trust} proposed Trust Score to calculate the agreement between the classifier and a modified nearest-neighbor classifier in the testing set. 
More recently, the idea of \textit{separate confidence estimator} was introduced by several works \cite{corbiere2019addressing,yu21bmvc}. Specifically, these works proposed to fix the task model, and instead conduct confidence estimation via a separate confidence estimator. Notably, Corbiere et al. \cite{corbiere2019addressing} proposed a separate confidence estimator called Confidnet and a new loss function called True Class Probability. Subsequently, Yu et al. \cite{yu21bmvc} proposed SLURP, a generic confidence estimator for regression tasks, that is specially targeted at task models that perform monocular depth estimation.  

In this paper, we also build a separate confidence estimator, since it has the benefit of not affecting the main task performance.
Different from previous works, we propose a novel meta-learning framework that simultaneously improves the performance of the confidence estimator under label imbalance and on out-of-distribution input data, in a unified manner.

\noindent\textbf{Label Imbalance in Confidence Estimation.}
Recently, using the the correctness of task model predictions ($C$) as labels, many existing confidence estimation methods \cite{hendrycks17baseline,gal2016dropout,corbiere2019addressing} have been shown to suffer from the label imbalance problem.
To solve this problem and enable the \textit{confidence estimator} to 
perform well under label imbalance, 
various methods have been proposed. Luo et al. \cite{luo2021learning} proposed a loss function called Steep Slope Loss to separate features w.r.t. correct and incorrect task model predictions from each other. 
Afterwards, Li et al. \cite{li2021identifying} proposed an extension to True Class Probability \cite{corbiere2019addressing} that uses a Distributional Focal Loss to focus more on predictions with higher uncertainty. 
Unlike previous methods that design strategies to handle a specific imbalanced distribution of correct and incorrect labels, we adopt a novel perspective, and tackle the label imbalance problem through meta-learning, which allows our confidence estimator to learn \textit{distribution-generalizable knowledge to tackle a variety of diverse label distributions}. This is done through construction of virtual testing sets that \textit{simulate various different label distributions}.

\noindent\textbf{Confidence Estimation on Out-of-distribution Data.}
As various distribution shifts exist between the training and testing data in real-world applications, the handling of out-of-distribution data inputs ($I$) is important for reliable confidence estimation.
To this end, Mukhoti et al. \cite{mukhoti2020calibrating} proposed to replace the cross entropy loss with the focal loss \cite{lin2017focal}, and utilize its implicit regularization effects to handle out-of-distribution data.
Tomani et al. \cite{tomani2021post} proposed to handle out-of-distribution data via applying perturbations on data from the validation set.
However, as these previous methods either emphasizes on the rare samples or fine-tunes on an additional set of samples, they can still be prone to overfit these rare samples or the additional set of samples.
Differently, in this work, we propose to use meta-learning and optimize the model through feedbacks from diverse virtual sets with diverse distributions. Thus, we can enable our model to learn knowledge that is more generalizable to various out-of-distribution data.

\noindent\textbf{Meta-learning.} 
MAML \cite{finn2017model}, a popular meta-learning method,
was originally designed to learn a good weight initialization that can quickly adapt to new tasks in testing,
which showed promise in few-shot learning. 
Subsequently, its extension \cite{li2018learning}, which requires no model updating on the unseen testing scenarios, has been applied beyond few-shot learning, to enhance model performance \cite{guo2020learning,bai2021person30k,huang2021metasets,xu2022meta}. 
Differently, we propose a novel framework via meta-learning to perform \textit{more reliable confidence estimation}. 
Through performing meta-learning on carefully constructed virtual training and virtual testing sets, 
we simultaneously improve the ability of our confidence estimator to generalize well to different distributions of $C$ and $I$.

\section{Method}

To conduct confidence estimation reliably, previous works have suggested two important qualities that a model should possess: 
the ability to \textit{perform well under label imbalance}, and the ability to \textit{handle various out-of-distribution data inputs}. 
We find that both qualities are actually acquired when the confidence estimator is trained to perform well across different distributions (w.r.t. either the correctness label $C$ or the data input $I$).
Hence, to train a more reliable confidence estimator, we leverage upon meta-learning that allows our confidence estimator to learn more distribution-generalizable knowledge to better tackle diverse distributions -- which is achieved by obtaining feedback from a \textit{virtual testing set} while concurrently updating using a \textit{virtual training set}.
A crucial part of our meta-learning algorithm is the virtual testing set construction, which needs to simulate diverse distributions to provide good feedback to the confidence estimator.
Specifically, at the start of each iteration, from the training set $D$, we first construct a virtual training set $D_{v\underline{~}tr}$ and a virtual testing set $D_{v\underline{~}te}$,
such that there are \textit{intentionally designed distribution differences between them}. 
To optimize the confidence estimator to possess both qualities discussed above, the virtual training set $D_{v\underline{~}tr}$ and the virtual testing set $D_{v\underline{~}te}$ are constructed to 
have different distributions of correctness labels $C$ every odd-numbered iteration and different distributions of data inputs $I$ every even-numbered iteration. 
After constructing $D_{v\underline{~}tr}$ and $D_{v\underline{~}te}$, our framework then uses them to train the confidence estimator through a \textit{virtual training and testing} procedure based on meta-learning.

Below, we first describe the \textit{virtual training and testing} scheme we use to train the confidence estimator in Sec.~\ref{sec:meta}. 
Next, in Sec.~\ref{sec:set}, we discuss how we construct our virtual training and virtual testing sets at the start of each iteration. 
Finally, we describe our framework as a whole in Sec.~\ref{sec:overall}.

\subsection{Virtual Training and Testing}
\label{sec:meta}

As mentioned above, at the start of each iteration, we first construct a virtual training set $D_{v\underline{~}tr}$ and a virtual testing set $D_{v\underline{~}te}$, such that there are intentionally designed distribution differences between them.
In this section, we assume that $D_{v\underline{~}tr}$ and $D_{v\underline{~}te}$ have been constructed, and describe how we utilize them via the \textit{virtual training and testing} scheme to train the confidence estimator to generalize to different distributions.

\begin{figure}[t]
\centering
\includegraphics[width=1.0\textwidth]{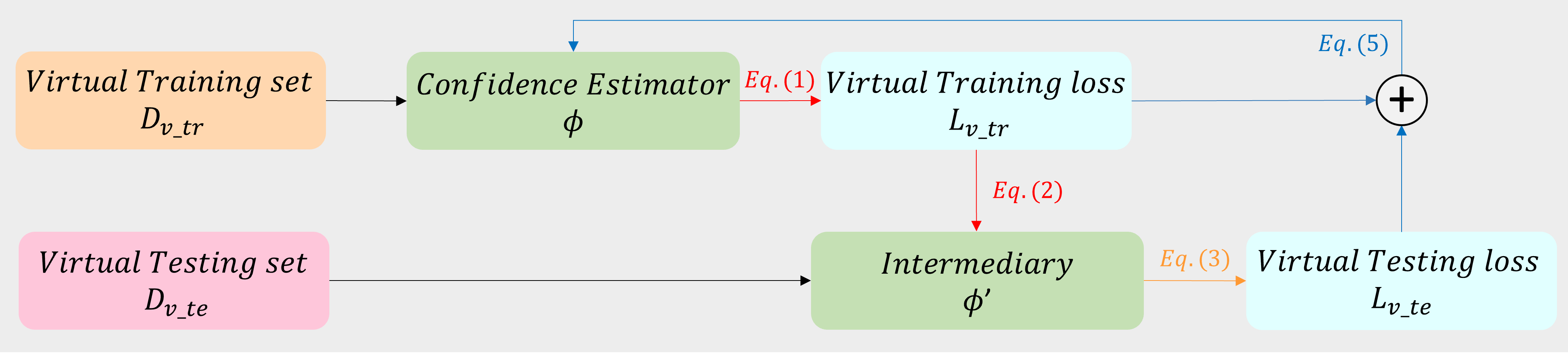}
\caption{
\textbf{Illustration of our virtual training and testing scheme.} 
(1) In the virtual training step, we conduct updates on the confidence estimator parameters $\phi$ with the virtual training set $D_{v\underline{~}tr}$, and obtain an intermediary $\phi'$ (which is indicated with red arrows).
(2) The intermediary $\phi'$ is then evaluated on the virtual testing set $D_{v\underline{~}te}$ with a different distribution from $D_{v\underline{~}tr}$ to obtain the virtual testing loss $L_{v\underline{~}te}$ (which is indicated with the yellow arrow). 
(3) Lastly, the virtual training loss $L_{v\underline{~}tr}$ and
the virtual testing loss $L_{v\underline{~}te}$ are used to update confidence estimator $\phi$ (indicated with blue arrows), such that it can generalize over diverse distributions and become more reliable. 
}
\label{fig:mech}
\end{figure}

Specifically, each iteration of the \textit{virtual training and testing} scheme contains three steps: 
(1) \textbf{Virtual training}. We first virtually train the confidence estimator using the virtual training set $D_{v\underline{~}tr}$ to simulate the conventional training procedure of the confidence estimator. 
(2) \textbf{Virtual testing}. After that, 
the confidence estimator is assessed (i.e., virtually tested) on the virtual testing set $D_{v\underline{~}te}$, which evaluates the performance on a different distribution from the virtual training set $D_{v\underline{~}tr}$.
(3) \textbf{Meta Optimization (Actual update)}. Finally, we incorporate the evaluation result (loss) calculated during virtual testing as a feedback to actually update the confidence estimator. 
This provides a feedback to the confidence estimator, 
that allows it to learn generalizable knowledge to tackle diverse distributions while training using the virtual training set $D_{v\underline{~}tr}$.
Below, we describe these three steps of the \textit{virtual training and testing} scheme in more detail. We also demonstrate these three steps in Fig.~\ref{fig:mech}. 

\textbf{Virtual Training.} 
During virtual training, we simulate the conventional training procedure of the confidence estimator, and first train the confidence estimator via gradient descent with data from the virtual training set $D_{v\underline{~}tr}$. 
Here we denote the confidence estimator parameters as $\phi$, the learning rate for virtual training as $\alpha$, and the loss function of the confidence estimator as $L$ (e.g., binary cross entropy loss). 
We can calculate the virtual training loss $L_{v\underline{~}tr}$ as:
\begin{equation}\label{eq:foundation_loss}
L_{v\underline{~}tr}(\phi) = L(\phi, D_{v\underline{~}tr})
\end{equation}
Using this loss, we can update our confidence estimator  parameters $\phi$ via gradient descent:
\begin{equation}\label{eq:foundation_gradient}
\phi’ = \phi - \alpha\nabla_\phi L_{v\underline{~}tr}(\phi)
\end{equation}
Note that we do not actually update the confidence estimator to be $\phi'$ (hence the term ``virtual"). 
Instead, the \textit{virtually trained} $\phi'$ is just an intermediary to calculate $L_{v\underline{~}te}$ in next step, and simulates what training on $D_{v\underline{~}tr}$ would be like.

\textbf{Virtual Testing.} 
In this step, we evaluate how the virtually updated confidence estimator $\phi'$ (that is trained on $D_{v\underline{~}tr}$) performs on the virtual testing set $D_{v\underline{~}te}$, which has a different distribution to $D_{v\underline{~}tr}$. 
\begin{equation}\label{eq:debias_gradient_1}
L_{v\underline{~}te}(\phi') = L(\phi', D_{v\underline{~}te})
\end{equation}
The computed virtual testing loss $L_{v\underline{~}te}$ measures the confidence estimator performance on $D_{v\underline{~}te}$, after one simulated training step on $D_{v\underline{~}tr}$, and can be used  to provide feedback on how we can update the confidence estimator parameters such that it can better generalize to different distributions (as is done in next step).

\textbf{Meta-optimization (Actual update).}
In the virtual training and virtual testing steps, we have computed the losses $L_{v\underline{~}tr}$ and $L_{v\underline{~}te}$ respectively.
In this step, we use them to optimize our confidence estimator to perform well on diverse distributions, by obtaining feedback from $L_{v\underline{~}te}$ while concurrently updating using $L_{v\underline{~}tr}$.
We first formulate our overall objective as:

\begin{equation}\label{eq:meta-optimization}
\begin{aligned}
     &\min_{\phi}\;\left\{L_{v\underline{~}tr}(\phi) + L_{v\underline{~}te}(\phi')\right\} \\
    =&\min_{\phi}\;\left\{L_{v\underline{~}tr}(\phi) + L_{v\underline{~}te}\big(\phi - \alpha\nabla_\phi L_{v\underline{~}tr}(\phi)\big)\right\}
\end{aligned}
\end{equation}

We highlight that, in Eq.~\ref{eq:meta-optimization}, our goal is to optimize $\phi$, 
and $\phi'$ is just used as a helpful intermediary in calculating $L_{v\underline{~}te}(\phi')$. 
After constructing our overall objective, we can then update $\phi$ via gradient descent for meta-optimization as:

\begin{equation}\label{eq:meta-update}
\begin{aligned}
    \phi \leftarrow \phi - \beta\nabla_\phi \Big (L_{v\underline{~}tr}(\phi) + L_{v\underline{~}te}\big(\phi - \alpha\nabla_\phi L_{v\underline{~}tr}(\phi)\big) \Big)
\end{aligned}
\end{equation}
where $\beta$ denotes the learning rate for meta-optimization. 
By updating the confidence estimator with the meta-optimization update rule in Eq.~\ref{eq:meta-update}, the confidence estimator is updated with knowledge that is more \textit{distribution-generalizable}, leading to a more reliable confidence estimator that is applicable to diverse distributions.
We explain this in more detail below.

During virtual training, we first update the confidence estimator $\phi$ to an intermediary $\phi'$ in Eq.~\ref{eq:foundation_gradient}.
In this step, 
the intermediary $\phi'$ can learn \textit{distribution-specific knowledge} (that is only specifically applicable to the distribution of $D_{v\underline{~}tr}$), as such knowledge can help improve  performance on $D_{v\underline{~}tr}$.
On the other hand, we note that for the intermediary (trained on $D_{v\underline{~}tr}$) to generalize well to the virtual testing set $D_{v\underline{~}te}$ in Eq.~\ref{eq:debias_gradient_1}
(which has a \textit{different distribution} compared to $D_{v\underline{~}tr}$ and where \textit{distribution-specific knowledge from $D_{v\underline{~}tr}$ does not apply}),
the intermediary needs to \textit{avoid learning distribution-specific knowledge} when learning on $D_{v\underline{~}tr}$, 
and instead \textit{learn more distribution-generalizable knowledge}.
This means that, 
the term $L_{v\underline{~}te}(\phi')$ in Eq.~\ref{eq:debias_gradient_1} provides a feedback which  
guides the learning towards acquiring more distribution-generalizable knowledge.

Importantly, based on our analysis above, 
as long as $D_{v\underline{~}tr}$ and $D_{v\underline{~}te}$ 
have \textit{different distributions and cannot be tackled with the same distribution-specific knowledge},
the confidence estimator will be encouraged to avoid learning distri-\\bution-specific knowledge, and focus on learning more distribution-generalizable knowledge in the meta-optimization step.
This also implies that, we do not aim to use the distribution of virtual testing set $D_{v\underline{~}te}$ to simulate the distribution of the real testing scenario (which is unknown during training) to learn distribution-generalizable knowledge that can tackle real testing scenarios.
We also present a more theoretical analysis of the efficacy of the meta-optimization rule in the supplementary material.

\subsection{Set Construction}
\label{sec:set}

In this section, we discuss how we construct a virtual training set $D_{v\underline{~}tr}$ and a virtual testing set $D_{v\underline{~}te}$ to have different distributions in each iteration of our virtual training and testing scheme (that is described in Sec.~\ref{sec:meta}). 
Specifically, \textit{at the start of each epoch}, we first split the training set $D$ into two halves: $D^C$ and $D^I$, which will be used to tackle the two different problems (w.r.t correctness labels $C$ and data inputs $I$).
Within the epoch, at the start of every odd-numbered iteration, we construct virtual training and testing sets from $D^C$ to tackle the label imbalance problem; on the other hand, at the start of every even-numbered iteration, we construct virtual training and testing sets from $D^I$ to tackle the out-of-distribution data input problem.

As there exist some differences between the distribution of $C$ and the distribution of $I$ (e.g., it is more difficult to characterize the distribution of data input $I$ and find input distributions that are different), 
we propose different set construction methods for each of them that provide diverse testing distributions in practice.
We highlight that, due to the \textit{unified nature of our framework},
tackling of these two different problems have now been reduced to a more straightforward designing of their respective set construction methods.
Below, we separately discuss each construction method.

\textbf{Constructing sets for correctness label $C$.}
With respect to the correctness label $C$, 
we construct virtual training and testing sets with different distributions in two steps.
\textbf{Step (C1)} \textit{At the start of each epoch}, we first randomly split $D^C$ into two subsets $D^C_1$ and $D^C_2$, 
where the first subset $D^C_1$ will be used to construct batches of $D_{v\underline{~}tr}$, and the second subset $D^C_2$ will be used to construct batches of $D_{v\underline{~}te}$. 
Then, we pre-compute the correctness label $C$ w.r.t. every sample in the second subset $D^C_2$ to facilitate $D_{v\underline{~}te}$ construction in that epoch. 
\textbf{Step (C2)} \textit{At the start of every odd-numbered iteration}, we randomly select a batch of data from the first subset $D^C_1$ to construct a virtual training set $D_{v\underline{~}tr}$. 
Next, we construct the virtual testing set $D_{v\underline{~}te}$ --
we want its distribution of $C$ to vary between iterations constantly, to simulate various distributions that are different from $D_{v\underline{~}tr}$.
Hence, we randomly select a percentage from $0$ to $100\%$ to set as the percentage of correct task model predictions (where $C=1$) in $D_{v\underline{~}te}$ each time.
Based on the sampled percentage of correct task model predictions, we randomly select a batch of samples from the second subset $D^C_2$ to construct the virtual testing set $D_{v\underline{~}te}$ to approximately match that percentage. 
This way, our virtual testing set $D_{v\underline{~}te}$ will have a different distribution of $C$ compared to $D_{v\underline{~}tr}$, with a high probability.

\textbf{Constructing sets for data input $I$.}
Besides, we also construct virtual training and testing sets to have different distributions w.r.t. the data input $I$. Note that, when using only a single dataset, constructing virtual training and testing sets to have different data input distributions is a difficult problem. 
Here we follow a simple and effective technique proposed in previous works \cite{huang2017arbitrary,matsuura2020domain,li2021feature} that can help to simulate a distribution shift within a dataset. 
Specifically, they found that the statistics (i.e., mean and standard deviation) computed spatially over the pixels within the convolutional feature map of an input image, are a \textit{compact representation that effectively captures the style and domain characteristics of this image}. 
Hence, we concatenate the convolutional feature statistics from all convolutional layers of our confidence estimator (into a single vector) as a representation of each input sample.
Then, following \cite{matsuura2020domain}, we use a K-means clustering technique to separate the convolutional feature statistics vectors of all the data in $D^I$ into different clusters, such that a data distribution shift is simulated between clusters, that will be used for constructing virtual training and testing sets with different distributions.

Specifically, our set construction for data input $I$ is done in two steps.
\textbf{Step (I1)} \textit{At the start of each epoch}, we first cluster $D^I$ into $N$ clusters by applying the K-means algorithm on the convolutional feature statistics vectors of samples in $D^I$. 
Among the $N$ clusters, we randomly select one cluster as $D^I_1$ that will be used to construct $D_{v\underline{~}tr}$ in this epoch. 
\textbf{Step (I2)} \textit{Then at the start of every even-numbered iteration}, we first randomly select a batch of data from the selected cluster $D^I_1$ to construct the virtual training set $D_{v\underline{~}tr}$. 
After that, we randomly select a cluster from the remaining $N-1$ clusters, and select a batch of data from this cluster to construct the virtual testing set $D_{v\underline{~}te}$.
For more details, please refer to the Supplementary.

After constructing virtual training and testing sets as discussed above, during experiments, we empirically observe consistent performance enhancement, as shown in Sec.~\ref{Sec:exp}, which shows the effectiveness of our set construction method.

\subsection{Overall Training and Testing Scheme}
\label{sec:overall}
In the above two sections, we have described the \textit{virtual training and testing} scheme and how we construct our virtual training and virtual testing sets. In this section, we summarize them and discuss the overall training and testing scheme of our framework. 
Specifically, in the training procedure of the confidence estimator, at the start of each iteration, we first construct the virtual training and testing sets to have different distributions (w.r.t correctness label $C$ in odd-numbered iterations and data input $I$ in even-numbered iterations) following Sec.~\ref{sec:set}. 
After that, the constructed virtual training and testing sets are used to train the confidence estimator through the \textit{virtual training and testing} scheme as discussed in Sec.~\ref{sec:meta}. 
Hence, we alternatingly deal with the label imbalance problem and the handling of out-of-distribution inputs over iterations, resulting a simulatenous tackling of both problems.
We demonstrate the overall training scheme of our framework in Alg.~\ref{algorithm_meta}. During testing, we follow the evaluation procedure of previous works \cite{yu21bmvc,corbiere2019addressing}.

\normalem
\begin{algorithm}[t]
\algsetup{linenosize=\tiny}
\scriptsize
 \nl Initialize $\phi$. \\
 \nl \For{$E$ epochs}{
 \nl Randomly split $D$ into two halves: $D^C$ and $D^I$. \\
 \nl Process $D^C$ and $D^I$ following Step (C1) and Step (I1) in Sec.~\ref{sec:set} respectively. \\
 \nl \For{$T$ iterations}{
  \nl  \uIf{$T$ is odd} {
    \nl Construct $D_{v\underline{~}tr}$ and $D_{v\underline{~}te}$ from $D^C$, following Step (C2) in Sec.~\ref{sec:set}. \\
  }
  \nl \Else{
    \nl Construct $D_{v\underline{~}tr}$ and $D_{v\underline{~}te}$ from $D^I$, following Step (I2) in Sec.~\ref{sec:set}. \\    
  }
  \nl Calculate the virtual training loss $L_{v\underline{~}tr}$ on $D_{v\underline{~}tr}$ using Eq.~\ref{eq:foundation_loss}: $L_{v\underline{~}tr}(\phi) = L(\phi, D_{v\underline{~}tr})$.\\
  \nl Calculate an updated version of confidence estimator ($\phi’$) using Eq.~\ref{eq:foundation_gradient}: $\phi’ = \phi - \alpha\nabla_\phi L_{v\underline{~}tr}(\phi)$. \\
  \nl Calculate the virtual testing loss $L_{v\underline{~}te}$ on $D_{v\underline{~}te}$ using Eq.~\ref{eq:debias_gradient_1}: $L_{v\underline{~}te}(\phi') = L(\phi', D_{v\underline{~}te})$.\\
  \nl Update using Eq.~\ref{eq:meta-update}:  $ \phi \leftarrow \phi - \beta\nabla_\phi \Big (L_{v\underline{~}tr}(\phi) + L_{v\underline{~}te}\big(\phi - \alpha\nabla_\phi L_{v\underline{~}tr}(\phi)\big) \Big)$.
 }
 }
 \scriptsize
 \caption{Overall Training Scheme}
 \label{algorithm_meta}
\end{algorithm}
\setlength{\textfloatsep}{6pt}
\section{Experiments}
\label{Sec:exp}

In this section, in order to verify the effectiveness of our proposed framework, we conduct experiments on various different tasks including monocular depth estimation and image classification. 
For monocular depth estimation, we only modify the training procedure by adding our framework and follow all the other experiment settings of \cite{yu21bmvc} for evaluation on various testing scenarios. 
For image classification, similarly, we merely change the training procedure to include our framework, and follow all the other experiment settings of \cite{corbiere2019addressing} to test our proposed method. 
We conduct all our experiments on an RTX 3090 GPU, and fix the task model during confidence estimator training.

\subsection{Confidence estimation on Monocular Depth Estimation}
\label{sec:dep}

\noindent\textbf{Settings and Implementation Details.} 
For monocular depth estimation, we follow \cite{yu21bmvc} and conduct two groups of experiments to evaluate our proposed framework. In the first experiment, we train our confidence estimator on KITTI

\begin{wraptable}[16]{r}{0.7\linewidth}
\caption{
Experiment results of confidence estimation on monocular depth estimation, with our model \textbf{trained on KITTI Eigen-split training set} following the setting in \cite{yu21bmvc}.
Our method performs the best across all metrics.
}
\label{Tab:KITTI}
\centering
\resizebox{0.7\textwidth}{!}
{\small
\begin{tabular}{ccccccc} \hline
\multirow{2}{*}{\textbf{Method}} &  \multicolumn{3}{c}{KITTI \cite{Geiger2013IJRR,uhrig2017sparsity,eigen2014depth}} & \multicolumn{3}{c}{CityScapes \cite{cordts2016cityscapes}} \\ \cmidrule(lr){2-4} \cmidrule(lr){5-7}
& \begin{tiny}\textbf{\makecell{AUSE-\\RMSE$\downarrow$}}\end{tiny} & \begin{tiny}\textbf{\makecell{AUSE-\\Absrel$\downarrow$}}\end{tiny} &
\begin{tiny}\textbf{\textbf{AUROC$\uparrow$}}\end{tiny}
& \begin{tiny}\textbf{\makecell{AUSE-\\RMSE$\downarrow$}}\end{tiny} & \begin{tiny}\textbf{\makecell{AUSE-\\Absrel$\downarrow$}}\end{tiny} &
\begin{tiny}\textbf{\textbf{AUROC$\uparrow$}}\end{tiny} \\
\hline\hline
MCDropout \cite{gal2016dropout} & 8.14 & 9.48 & 0.686 & 9.42 & 9.52 & 0.420\\
Empirical Ensembles & 3.17 & 5.02 & 0.882 & 11.56 & 13.14 & 0.504\\
Single PU \cite{kendall2017uncertainties} & 1.89 & 4.59 & 0.882 & 9.91 & 9.96 & 0.386\\
Deep Ensembles \cite{lakshminarayanan2017simple} & 1.68 & 4.32 & 0.897 & 11.47 & 9.36 & 0.501\\
True Class Probability \cite{corbiere2019addressing} & 1.76 & 4.24 & 0.892 & 10.48 & 5.75 & 0.519\\ 
SLURP \cite{yu21bmvc} & 1.68 & 4.36 & 0.895 & 9.48 & 10.90 & 0.400\\
SLURP + Reweight & 1.67 & 4.29 & 0.896 & 9.39 & 10.41 & 0.402 \\
SLURP + Resample \cite{burnaev2015influence} & 1.67 & 4.28 & 0.896 & 9.37 & 10.35 & 0.404 \\
SLURP + Dropout \cite{srivastava2014dropout} & 1.67 & 4.20 & 0.896 & 9.29 & 10.01 & 0.412\\
SLURP + Focal loss \cite{lin2017focal} & 1.67 & 4.18 & 0.895 & 9.30 & 10.14 & 0.410 \\
SLURP + Mixup \cite{zhang2018mixup} & 1.67 & 4.15 & 0.896 & 9.17 & 10.01 & 0.420 \\
SLURP + Resampling + Mixup & 1.67 & 4.07 & 0.896 & 8.99 & 9.64 & 0.431 \\
\hline
SLURP + Ours(\textit{tackling label imbalance only}) & 1.66 & 3.84 & 0.897 & 8.75 & 7.79 & 0.509 \\
SLURP + Ours(\textit{tackling out-of-distribution inputs only}) & 1.66 & 3.90 & 0.897 & 8.54 & 6.90 & 0.524 \\
\textbf{SLURP + Ours(full)} & \textbf{1.65} & \textbf{3.62} & \textbf{0.898} & \textbf{8.26} & \textbf{5.32} & \textbf{0.601}\\
\hline
\end{tabular}}
\end{wraptable}

\noindent Eigen-split training set \cite{eigen2014depth,Geiger2013IJRR,uhrig2017sparsity}, and evaluate the trained confidence estimator on two testing scenarios: KITTI Eigen-split testing set from the \textit{same dataset}, 
and Cityscapes \cite{cordts2016cityscapes} testing set from a \textit{different dataset}. 

In the second experiment, 
we further evaluate our framework under \textit{different weather conditions}.
We fine tune our trained confidence estimator on Cityscapes training set, and evaluate it on several testing scenarios: Cityscapes testing set, Foggy Cityscapes-DBF \cite{sakaridis2018model} testing set with three severity levels, and Rainy Cityscapes \cite{hu2019depth} testing set with three severity levels.

We emphasize that in these experiments, the distribution of $I$ will face a large shift from training conditions due to the cross-dataset/cross-weather setting. Moreover, there is also obvious imbalance in the distribution of $C$.
Specifically, \textbf{the distribution of correct and incorrect labels of C in the KITTI Eigen-split training set is quite imbalanced (99.8\%:0.2\%)}. This means, there are both obvious label ($C$) imbalance problem and input ($I$) out-of-distribution problem.

In both above-mentioned experiments, we use the same backbone as SLURP \cite{yu21bmvc}, which is described in more detail in the supplementary material. 
Following the setting in \cite{yu21bmvc}, we use the area under sparsification error corresponding to square error (\textbf{AUSE-RMSE}), the area under sparsification error corresponding to absolute relative error (\textbf{AUSE-Absrel}) \cite{eigen2014depth}, and the area under the receiver operating characteristic (\textbf{AUROC}) as our evaluation metrics for the confidence estimator.
We also follow \cite{yu21bmvc} to regard the depth prediction of \textit{each single pixel} to be correct ($C=1$) if the relative difference between the depth prediction and the ground truth is less than $25\%$.
Correspondingly, we also regard the depth prediction of \textit{an input image} to be correct ($C=1$) if the average relative difference among all its pixels (w.r.t the ground truth image) is less than $25\%$.

At the start of every training epoch, we randomly select $60\%$ of data from $D^C$ to construct the first subset $D^C_1$, and use the remaining as the second subset $D^C_2$. 
On the other hand, $D^I$ is clustered into 6 clusters (i.e., $N=6$), and one cluster is randomly selected to be $D^I_1$. We ablate these decisions in Sec.~\ref{sec:ablation}.
During training, we set the learning rate ($\alpha$) for \textit{virtual training} to $5e-4$, and the learning rate ($\beta$) for meta-optimization to $1e-4$.

\begin{table}[t]
\caption{
Experiment results of confidence estimation on monocular depth estimation, with our model \textbf{fine-tuned on CityScapes \cite{cordts2016cityscapes}} following the setting in \cite{yu21bmvc}. In this table, $s$ indicates severity. The higher $s$ is, more severe the rain or the fog is.
Our method performs the best across all metrics and testing scenarios.
}
\label{Tab:cityscape}
\resizebox{\textwidth}{!}
{\small
\begin{tabular}{cccccccccccccccccccccc} \hline
\multirow{2}{*}{\textbf{Method}} 
& \multicolumn{3}{c}{CityScapes \cite{cordts2016cityscapes}} 
& \multicolumn{3}{c}{\makecell{CityScapes\\Foggy s = 1 \cite{sakaridis2018model}}} 
& \multicolumn{3}{c}{\makecell{CityScapes\\Foggy s = 2 \cite{sakaridis2018model}}} 
& \multicolumn{3}{c}{\makecell{CityScapes\\Foggy s = 3 \cite{sakaridis2018model}}} 
& \multicolumn{3}{c}{\makecell{CityScapes\\Rainy s = 1 \cite{hu2019depth}}}
& \multicolumn{3}{c}{\makecell{CityScapes\\Rainy s = 2 \cite{hu2019depth}}}
& \multicolumn{3}{c}{\makecell{CityScapes\\Rainy s = 3 \cite{hu2019depth}}}\\ 
\cmidrule(lr){2-4} \cmidrule(lr){5-7} \cmidrule(lr){8-10} \cmidrule(lr){11-13} \cmidrule(lr){14-16} \cmidrule(lr){17-19} \cmidrule(lr){20-22}
& \begin{tiny}\textbf{\makecell{AUSE-\\RMSE$\downarrow$}}\end{tiny} & \begin{tiny}\textbf{\makecell{AUSE-\\Absrel$\downarrow$}}\end{tiny} &
\begin{tiny}\textbf{\textbf{AUROC$\uparrow$}}\end{tiny}
& \begin{tiny}\textbf{\makecell{AUSE-\\RMSE$\downarrow$}}\end{tiny} & \begin{tiny}\textbf{\makecell{AUSE-\\Absrel$\downarrow$}}\end{tiny} &
\begin{tiny}\textbf{\textbf{AUROC$\uparrow$}}\end{tiny}
& \begin{tiny}\textbf{\makecell{AUSE-\\RMSE$\downarrow$}}\end{tiny} & \begin{tiny}\textbf{\makecell{AUSE-\\Absrel$\downarrow$}}\end{tiny} &
\begin{tiny}\textbf{\textbf{AUROC$\uparrow$}}\end{tiny}
& \begin{tiny}\textbf{\makecell{AUSE-\\RMSE$\downarrow$}}\end{tiny} & \begin{tiny}\textbf{\makecell{AUSE-\\Absrel$\downarrow$}}\end{tiny} &
\begin{tiny}\textbf{\textbf{AUROC$\uparrow$}}\end{tiny}
& \begin{tiny}\textbf{\makecell{AUSE-\\RMSE$\downarrow$}}\end{tiny} & \begin{tiny}\textbf{\makecell{AUSE-\\Absrel$\downarrow$}}\end{tiny} &
\begin{tiny}\textbf{\textbf{AUROC$\uparrow$}}\end{tiny}
& \begin{tiny}\textbf{\makecell{AUSE-\\RMSE$\downarrow$}}\end{tiny} & \begin{tiny}\textbf{\makecell{AUSE-\\Absrel$\downarrow$}}\end{tiny} &
\begin{tiny}\textbf{\textbf{AUROC$\uparrow$}}\end{tiny}
& \begin{tiny}\textbf{\makecell{AUSE-\\RMSE$\downarrow$}}\end{tiny} & \begin{tiny}\textbf{\makecell{AUSE-\\Absrel$\downarrow$}}\end{tiny} &
\begin{tiny}\textbf{\textbf{AUROC$\uparrow$}}\end{tiny}
\\ \hline\hline
MCDropout \cite{gal2016dropout} 
& 7.72 & 8.13 & 0.705 & 7.06 & 8.73 & 0.659 
& 7.14 & 8.36 & 0.667 & 7.30 & 8.27 & 0.665
& 7.80 & 8.36 & 0.700 & 7.82 & 8.20 & 0.704
& 7.84 & 7.87 & 0.715 \\
Empirical Ensembles 
& 8.20 & 7.50 & 0.786 & 7.29 & 6.92 & 0.757 
& 6.90 & 6.48 & 0.767 & 6.66 & 6.03 & 0.778
& 7.82 & 7.33 & 0.783 & 7.53 & 7.09 & 0.791
& 7.28 & 6.80 & 0.801 \\
Single PU \cite{kendall2017uncertainties} 
& 4.35 & 6.44 & 0.741 & 4.17 & 6.55 & 0.731 
& 4.27 & 6.79 & 0.731 & 4.35 & 6.44 & 0.742
& 3.42 & 6.78 & 0.842 & 3.42 & 6.55 & 0.847
& 3.48 & 6.19 & 0.851 \\
Deep Ensembles \cite{lakshminarayanan2017simple} 
& 3.03 & 6.81 & 0.856 & 3.42 & 6.68 & 0.746 
& 3.35 & 6.24 & 0.756 & 3.28 & 5.85 & 0.767
& 3.05 & 6.58 & 0.852 & 2.98 & 6.35 & 0.857
& 2.93 & 6.01 & 0.863 \\
True Class Probability \cite{corbiere2019addressing} 
& 4.05 & 6.34 & 0.821 & 4.89 & 7.26 & 0.697 
& 4.68 & 6.86 & 0.714 & 4.59 & 6.64 & 0.729
& 3.98 & 6.21 & 0.824 & 3.86 & 6.02 & 0.833
& 3.70 & 5.78 & 0.846 \\
SLURP \cite{yu21bmvc} 
& 3.05 & 6.55 & 0.849 & 3.39 & 5.62 & 0.788 
& 3.36 & 5.28 & 0.794 & 3.41 & 5.05 & 0.801
& 3.04 & 6.25 & 0.847 & 3.01 & 6.06 & 0.852
& 3.08 & 5.80 & 0.857 \\
SLURP + Reweight & 2.56 & 5.47 & 0.861 & 2.71 & 5.14 & 0.804 & 2.89 & 5.06 & 0.811 & 2.93 & 4.46 & 0.819 & 2.85 & 6.07 & 0.854 & 2.89 & 5.87 & 0.868 & 2.45 & 5.14 & 0.864 \\
SLURP + Resample \cite{burnaev2015influence} & 2.51 & 5.32 & 0.865 & 2.69 & 5.11 & 0.805 & 2.76 & 4.99 & 0.813 & 2.89 & 4.27 & 0.823 & 2.72 & 5.99 & 0.857 & 2.77 & 5.84 & 0.871 & 2.31 & 4.85 & 0.866 \\
SLURP + Dropout \cite{srivastava2014dropout} & 2.88 & 5.39 & 0.857 & 2.52 & 4.91 & 0.813 & 2.55 & 4.78 & 0.819 & 2.74 & 4.01 & 0.835 & 2.56 & 5.68 & 0.863 & 2.65 & 5.47 & 0.880 & 2.27 & 4.77 & 0.870 \\
SLURP + Focal loss \cite{lin2017focal} & 2.75 & 5.20 & 0.859 & 2.49 & 4.87 & 0.816 & 2.47 & 4.69 & 0.825 & 2.67 & 4.09 & 0.830 & 2.49 & 5.55 & 0.867 & 2.41 & 5.23 & 0.884 & 2.18 & 4.81 & 0.867 \\
SLURP + Mixup \cite{zhang2018mixup} & 2.49 & 5.13 & 0.866 & 2.28 & 4.59 & 0.827 & 2.33 & 4.41 & 0.830 & 2.44 & 3.92 & 0.847 & 2.41 & 5.44 & 0.869 & 2.34 & 5.11 & 0.889 & 2.03 & 4.51 & 0.872 \\
SLURP + Resample + Mixup & 2.31 & 4.97 & 0.869 & 2.05 & 4.29 & 0.836 & 2.08 & 4.24 & 0.842 & 2.29 & 3.79 & 0.853 & 2.33 & 5.14 & 0.880 & 2.09 & 4.98 & 0.877 & 1.94 & 3.77 & 0.877 \\
\hline
SLURP + Ours(\textit{tackling label imbalance only}) & 1.33 & 1.76 & 0.908 
& 1.84 & 2.64 & 0.874 & 1.90 & 2.51 & 0.871 
& 1.98 & 2.34 & 0.864
& 1.81 & 3.17 & 0.889 & 1.77 & 3.10 & 0.891
& 1.71 & 2.55 & 0.889  \\
 
SLURP + Ours(\textit{tackling out-of-distribution inputs only}) & 1.42 & 1.88 & 0.900 
& 1.61 & 2.19 & 0.890 & 1.66 & 2.20 & 0.891 
& 1.76 & 2.09 & 0.881 
& 1.59 & 2.78 & 0.901 & 1.60 & 2.75 & 0.904 
& 1.63 & 1.97 & 0.890 \\
\textbf{SLURP + Ours(full)} 
& \textbf{0.60} & \textbf{0.62} & \textbf{0.933} & \textbf{0.73} & \textbf{0.63} & \textbf{0.934}
& \textbf{0.80} & \textbf{0.58} & \textbf{0.937}
& \textbf{0.93} & \textbf{0.58} & \textbf{0.938} & \textbf{0.85} & \textbf{0.69} & \textbf{0.923} & \textbf{0.96} & \textbf{0.68} & \textbf{0.925} & \textbf{1.08} & \textbf{0.80} & \textbf{0.909}\\
\hline
\end{tabular}}
\end{table}

\noindent\textbf{Experiment Results.} 
In both experiments as shown in Tab.~\ref{Tab:KITTI} and Tab.~\ref{Tab:cityscape}, we compare our framework with both existing confidence estimation methods and other representative methods on tackling label imbalance problem (Reweight and Resample \cite{burnaev2015influence}) and improving out-of-distribution data generalization (Dropout \cite{srivastava2014dropout}, Focal loss \cite{lin2017focal}, and Mixup \cite{zhang2018mixup}). Besides, we also compare with the combination of Resample \cite{burnaev2015influence} and Mixup \cite{zhang2018mixup}, which have been shown to effectively tackle label imbalance problem and improving out-of-distribution data generalization respectively.
We reimplement all these methods on the same backbone (densenet161) as SLURP and ours.
Furthermore, to better assess the effectiveness of our framework in tackling either of the two problems individually,
we also assess the following two variants of our framework, i.e., the variant (\textbf{tackling label imbalance only}) that only constructs virtual training and testing sets w.r.t. distribution of the correctness label $C$, and the variant (\textbf{tackling out-of-distribution inputs only}) that only constructs virtual training and testing sets w.r.t. distribution of the data input $I$.

As shown in Tab.~\ref{Tab:KITTI}, 
as compared to all other methods,
our confidence estimator that is trained on KITTI Eigen-split training set achieves the \textit{best performance across all metrics and testing scenarios},
including both the KITTI Eigen-split testing set from the same dataset
and Cityscapes testing set from a different dataset.
This demonstrates that, by training the confidence estimator towards improvements on both qualities with our framework, the confidence estimator can become more reliable.
Besides, as shown in Tab.~\ref{Tab:KITTI}, both variants of our framework improve the performance of the SLURP baseline, demonstrating that both individual set construction methods
can lead to a more reliable confidence estimator, through our \textit{virtual training and testing} scheme.

In Tab.~\ref{Tab:cityscape}, we evaluate the reliability of our confidence estimator under different weather conditions, and report results on the testing sets of Cityscapes,
Foggy Cityscapes-DBF and Rainy Cityscapes, where three different weather severity levels are reported for the latter two datasets.
We highlight that, under different weather conditions with different severity levels, there are fluctuating degrees of distribution shifts (of $C$ and $I$) as compared to the Cityscapes training set that are challenging for a confidence estimator to handle.
Our framework \textit{outperforms all other methods on all reported metrics}, demonstrating that our framework effectively leads the confidence estimator to learn knowledge that can generalize to diverse distributions.

\subsection{Confidence estimation on Image Classification}

\noindent\textbf{Settings and Implementation Details.} To evaluate our proposed framework on image classification, we
follow previous works \cite{jiang2018trust,corbiere2019addressing} and conduct experiments on both MNIST \cite{lecun1998gradient} and CIFAR-10 \cite{krizhevsky2009learning}. On MNIST, the confidence estimator is trained on MNIST training set and tested on MNIST testing set; and on CIFAR-10, the confidence estimator is trained on CIFAR-10 training set and tested on CIFAR-10 testing set. 
Note that, in these experiments, even though training/testing are done on same dataset, there are still data distribution shift issues, as shown in \cite{rabanser2019failing,matsuura2020domain}.

On both datasets, we use the same backbone (Confidnet) as TCP \cite{corbiere2019addressing}, which is described in more detail in the supplementary material. 
Following \cite{hendrycks17baseline,corbiere2019addressing}, we 
report scores on 4 evaluation metrics:
the False positive rate at $95\%$ True positive rate (\textbf{FPR-95\%-TPR}), 
the area under the precision-recall curve with respect to $C=0$ labels (\textbf{AUPR-Error}), the area under the precision-recall curve with respect to $C=1$ labels (\textbf{AUPR-Success}),
and the area under the receiver operating characteristic (\textbf{AUROC}).

Our set construction hyperparameters are set similarly to the monocular depth estimation experiments.
At the start of every training epoch, we randomly select $60\%$ of data from $D^C$ to construct $D^C_1$ and designate the rest as $D^C_2$, while $D^I$ is split into 6 clusters (i.e., $N=6$).
On MNIST, we set the learning rate ($\alpha$) for \textit{virtual training} to $1e-4$, and the learning rate ($\beta$) for meta-optimization to $1e-4$. On CIFAR-10, we set the learning rate ($\alpha$) for \textit{virtual training} to $1e-5$, and the learning rate ($\beta$) for meta-optimization to $1e-5$.

\begin{wraptable}[22]{r}{0.7\linewidth}
	\caption{Results on MNIST and CIFAR-10 using same backbone as \cite{corbiere2019addressing}.
	When using our variant that tackles label imbalance only, we obtain obvious improvements in the AUPR-Error metric, showing efficacy on tackling cases where $C=0$. 
	}
	\label{Tab:classification}
	\centering
	\resizebox{0.65\textwidth}{!}{
	\begin{tabular}{cccccc}
		\hline
\textbf{Dataset} & \textbf{Method}
& \begin{tiny}\textbf{\makecell{FPR-95\%\\-TPR$\downarrow$}}\end{tiny} 
	 & \begin{tiny}\textbf{\makecell{AUSE-\\Error$\uparrow$}}\end{tiny}
	 & \begin{tiny}\textbf{\makecell{AUSE-\\Success$\uparrow$}}\end{tiny} &
     \begin{tiny}\textbf{\textbf{AUROC$\uparrow$}}\end{tiny}\\
		\hline \hline
		\multirow{15}{*}{MNIST \cite{lecun1998gradient}} 
		& Maximum Class Probability \cite{hendrycks17baseline}  & 5.56 & 35.05 & \textbf{99.99} & 98.63\\
		& MCDropout \cite{gal2016dropout}  & 5.26 & 38.50 & \textbf{99.99} & 98.65\\
		& Trust Score \cite{jiang2018trust}  & 10.00 & 35.88 & 99.98 & 98.20\\
		& Steep Slope Loss \cite{luo2021learning}  & \textbf{2.22} & 40.86 & \textbf{99.99} & 98.83\\
		& True Class Probability (TCP) \cite{corbiere2019addressing}  & 3.33 & 45.89 & \textbf{99.99} & 98.82\\
		& Balancecd TCP \cite{li2021identifying}  & 4.44 & 43.03 & \textbf{99.99} & 98.67 \\
		& TCP + Reweight & 7.78 & 31.67 & 99.98 & 98.06\\
        & TCP + Resample \cite{burnaev2015influence} & 6.67 & 33.57 & 99.98 & 98.32\\
        & TCP + Dropout \cite{srivastava2014dropout} & 3.33 & 43.05 & \textbf{99.99} & 98.79 \\
        & TCP + Focal loss \cite{lin2017focal} & 4.44 & 42.65 & \textbf{99.99} & 98.73 \\
        & TCP + Mixup \cite{zhang2018mixup} & 4.44 & 45.73 & \textbf{99.99} & 98.80 \\
        & TCP + Resample + Mixup & 4.44 & 45.45 & \textbf{99.99} & 98.78 \\
        \cline{2-6}
        & TCP + Ours(\textit{tackling label imbalance only}) & \textbf{2.22} & 46.71 & \textbf{99.99} & 98.87 \\
        & TCP + Ours(\textit{tackling out-of-distribution inputs only}) & 3.33 & 45.91 & \textbf{99.99} & 98.85 \\
		& \textbf{TCP + Ours(full)}  & \textbf{2.22} & \textbf{47.05} & \textbf{99.99} & \textbf{98.91}\\
        \hline
        \multirow{15}{*}{CIFAR-10 \cite{krizhevsky2009learning}} & Maximum Class Probability \cite{hendrycks17baseline}  & 47.50 & 45.36 & 99.19 & 91.53\\
		& MCDropout \cite{gal2016dropout}  & 49.02 & 46.40 & \textbf{99.27} & 92.08\\
		& Trust Score \cite{jiang2018trust}  & 55.70 & 38.10 & 98.76 & 88.47\\
		& Steep Slope Loss \cite{luo2021learning}  & \textbf{44.69 }& 50.28  & 99.26 & 92.22\\
        & True Class Probability (TCP)\cite{corbiere2019addressing}  & 44.94 & 49.94 & 99.24 & 92.12\\
        & Balancecd TCP \cite{li2021identifying}  & 45.33 & 49.79 & 99.25 & 92.19\\
        & TCP + Reweight & 45.20 & 49.77 & 99.25 & 92.18 \\
        & TCP + Resample \cite{burnaev2015influence} & 45.71 & 49.81 & 99.25 & 92.20  \\
        & TCP + Dropout \cite{srivastava2014dropout} & 45.45 & 49.63 & 99.25 & 92.19 \\
        & TCP + Focal loss \cite{lin2017focal} & 45.07 & 49.46 & 99.24 & 92.09 \\
        & TCP + Mixup \cite{zhang2018mixup} & 45.33 & 49.68 & 99.25 & 92.18 \\
        & TCP + Resample + Mixup & 45.20 & 49.66 & 99.25 & 92.18 \\
        \cline{2-6}
        & TCP + Ours(\textit{tackling label imbalance only}) & 44.81 & 50.27 & 99.26 & 92.23 \\
        & TCP + Ours(\textit{tackling out-of-distribution inputs only}) 
        & 44.81 & 50.26 & 99.26 & 92.23 \\
		& \textbf{TCP + Ours(full)}  & \textbf{44.69} & \textbf{50.30} & \textbf{99.27} & \textbf{92.26}\\
		\hline
	\end{tabular}}
\end{wraptable}

\noindent\textbf{Results and Analysis.} In Tab.~\ref{Tab:classification} we report results using our framework, as well as existing image classification confidence estimation methods and various other representative methods introduced previously
(i.e., Reweight and Resample \cite{burnaev2015influence}, Dropout \cite{srivastava2014dropout}, Focal loss \cite{lin2017focal}, Mixup \cite{zhang2018mixup}, and Resample \cite{burnaev2015influence} + Mixup \cite{zhang2018mixup}).
As shown in Tab.~\ref{Tab:classification}, on both the MNIST testing set and the CIFAR-10 testing set, our framework achieves the \textit{best performance across all metrics}, which demonstrates that our framework can improve confidence estimator performance effectively. 
The variants of our framework also achieves improvements, which shows the superiority of our framework both in tackling label imbalance problem and improving generalization to distribution shifts in input data.
In particular, the variant tackling the label imbalance problem achieves an obvious improvement gain on the AUPR-Error metric which focuses on the performance where $C=0$.

\noindent\textbf{Experiments on Imagenet.}
For confidence estimation on image classification, besides evaluating our framework on small scale datasets including MNIST \cite{lecun1998gradient} and CIFAR-10  \cite{krizhevsky2009learning} following many previous works \cite{jiang2018trust,corbiere2019addressing}, we also evaluate our framework on the large-scale dataset Imagenet \cite{deng2009imagenet} following \cite{luo2021learning}. Here, we use the same backbone as \cite{luo2021learning}. As shown in Tab.~\ref{Tab:imagenet}, after incorporating our framework, we observe a significant performance improvement, which further shows the effectiveness of our method in a large-scale scenario with more classes and larger images, which is more realistic.

\subsection{Additional Ablation Studies}
\label{sec:ablation}

In this section and in the supplementary material, we conduct more extensive ablation studies on the monocular depth estimation task, with a confidence estimator that is fine-tuned on CityScapes training set. 
Specifically, our framework is evaluated on the CityScapes testing set, as well as both the Foggy Cityscapes-DBF testing set and Rainy Cityscapes testing set with the highest severity level (i.e., s = 3).

\noindent\textbf{Impact of second-order gradient.}
In our framework, we update the confidence estimator utilizing the \textit{virtual training and testing} scheme through a second-order gradient $\nabla_\phi \Big (L_{v\underline{~}tr}(\phi) + L_{v\underline{~}te}\big(\phi - \alpha\nabla_\phi L_{v\underline{~}tr}(\phi)\big) \Big)$. 
To investigate the impact of such a second-order gradient, we compare our framework (\textbf{meta-learning scheme}) with a variant (\textbf{joint-training scheme}) that still constructs virtual training and testing sets in the same way, but optimizes the confidence estimator through $\nabla_\phi \big (L_{v\underline{~}tr}(\phi) + L_{v\underline{~}te}(\phi)\big)$ without utilizing the \textit{virtual training and testing} scheme. 
As shown in Tab.~\ref{Tab:ablation_study_4}, our framework consistently outperforms this variant, which shows effectiveness of the \textit{virtual training and testing} scheme.

\begin{table}[t]
\parbox{0.42\textwidth}{
	\caption{
	Experiment results on Imagenet \cite{deng2009imagenet}, where our framework is applied on Steep Slope Loss \cite{luo2021learning}, which is the current state-of-the-art. We obtain a significant performance improvement.
	}
	\label{Tab:imagenet}
	\resizebox{0.42\textwidth}{!}{
	\begin{tabular}{ccccc}
		\hline
	 \textbf{Method} 
	 & \begin{tiny}\textbf{\makecell{FPR-95\%\\-TPR$\downarrow$}}\end{tiny} 
	 & \begin{tiny}\textbf{\makecell{AUSE-\\Error$\uparrow$}}\end{tiny}
	 & \begin{tiny}\textbf{\makecell{AUSE-\\Success$\uparrow$}}\end{tiny} &
     \begin{tiny}\textbf{\textbf{AUROC$\uparrow$}}\end{tiny}
	 \\
		\hline \hline
		 Steep Slope Loss \cite{luo2021learning} & 80.48 & 10.26 & 93.01 & 73.68 \\
		\textbf{Steep Slope Loss + Ours(full)} & \textbf{76.70} & \textbf{10.33} & \textbf{94.11} & \textbf{78.60}\\
		\hline
	\end{tabular}}
	}
\hspace{0.01\textwidth}
\parbox{0.56\textwidth}{
\caption{
Ablation studies conducted on the effectiveness of the virtual training and testing scheme.
}
\label{Tab:ablation_study_4}
\resizebox{0.56\textwidth}{!}
{\small
\begin{tabular}{cccccccccc} \hline
\multirow{2}{*}{\textbf{Method}} 
& \multicolumn{3}{c}{CityScapes \cite{cordts2016cityscapes}} 
& \multicolumn{3}{c}{\makecell{CityScapes\\Foggy s = 3 \cite{sakaridis2018model}}} 
& \multicolumn{3}{c}{\makecell{CityScapes\\Rainy s = 3 \cite{hu2019depth}}}\\ 
\cmidrule(lr){2-4} \cmidrule(lr){5-7} \cmidrule(lr){8-10} 
& \begin{tiny}\textbf{\makecell{AUSE-\\RMSE$\downarrow$}}\end{tiny} & \begin{tiny}\textbf{\makecell{AUSE-\\Absrel$\downarrow$}}\end{tiny} &
\begin{tiny}\textbf{\textbf{AUROC$\uparrow$}}\end{tiny}
& \begin{tiny}\textbf{\makecell{AUSE-\\RMSE$\downarrow$}}\end{tiny} & \begin{tiny}\textbf{\makecell{AUSE-\\Absrel$\downarrow$}}\end{tiny} &
\begin{tiny}\textbf{\textbf{AUROC$\uparrow$}}\end{tiny}
& \begin{tiny}\textbf{\makecell{AUSE-\\RMSE$\downarrow$}}\end{tiny} & \begin{tiny}\textbf{\makecell{AUSE-\\Absrel$\downarrow$}}\end{tiny} &
\begin{tiny}\textbf{\textbf{AUROC$\uparrow$}}\end{tiny}
\\ \hline\hline
Baseline(SLURP)
& 3.05 & 6.55 & 0.849 & 3.41 & 5.05 & 0.801
& 3.08 & 5.80 & 0.857 \\
\hline
Joint-training scheme & 2.47 & 5.11 & 0.867 & 2.63 & 4.01 & 0.829 & 2.12 & 3.98 & 0.869 \\
\hline
Meta-learning scheme
& 0.60 & 0.62 & 0.933 
& 0.93 & 0.58 & 0.938 & 1.08 & 0.80 & 0.909\\
\hline
\end{tabular}}}

\end{table}

\section{Conclusion}

In this paper, we propose a unified framework that improves the reliability of confidence estimators, through simultaneously improving their performance
under label imbalance and their handling of various out-of-distribution data inputs.
Through carefully constructing virtual training and testing sets with different distributions w.r.t. both the correctness label $C$ and the data input $I$, our framework trains the confidence estimator with a \textit{virtual training and testing} scheme and leads it to learn knowledge that is more generalizable to different distributions (w.r.t. both the $C$ and $I$). 
To validate the general effectiveness of our framework, we apply our framework to confidence estimation methods on both monocular depth estimation and image classification tasks, and show consistent improvements on both.

\section*{Acknowledgement}
This work is supported by National Research Foundation, Singapore under its AI Singapore Programme (AISG Award No: AISG-100E-2020-065), Ministry of Education Tier 1 Grant and SUTD Startup Research Grant.
\clearpage
%
%
\bibliographystyle{splncs04}
\bibliography{egbib}
\end{document}